\documentclass{article}

\makeatletter
\renewcommand{\@maketitle}{%
  \newpage
  \null
  \vskip 2em%
  \begin{center}%
    \let \footnote \thanks
    {\LARGE \@title \par}%
    \vskip 1.5em%
    {\large \lineskip 0.5em%
      \begin{tabular}[t]{c}%
        \@author
      \end{tabular}\par}%
    \vskip 1em%
    {\large \@date}%
  \end{center}%
  \par
  \vskip 1.5em}
\makeatother
\usepackage{PRIMEarxiv}

\usepackage[utf8]{inputenc} 
\usepackage[T1]{fontenc}    
\usepackage{hyperref}       
\usepackage{url}            
\usepackage{booktabs}       
\usepackage{amsfonts}       
\usepackage{nicefrac}       
\usepackage{microtype}      
\usepackage{lipsum}
\usepackage{fancyhdr}       
\usepackage{graphicx}       
\graphicspath{{media/}}     

\title{Student Classroom Behavior Recognition Based on Improved YOLOv8s
\thanks{\textit{\underline{Citation}}: 
\textbf{Authors. Title. Pages.... DOI:000000/11111.}} 
}

\author{
  Xiang Gao, Shuai Hang \\
  School of Physics and Information Technology  \\
  Shaanxi Normal University \\
  Xi’an, 710062, China \\
  \texttt{\{Xiang Gao\}gaoxiang.gnaixoag@gmail.com} \\
}

\begin{document}
\maketitle

\begin{abstract}
In classroom teaching, student behavior can reflect their learning state and classroom participation, which is of great significance for teaching quality analysis. To address the problems of dense student targets, numerous small objects, frequent occlusions, and imbalanced class distribution in real classroom scenes, this paper proposes an improved student classroom behavior recognition model named ALC-YOLOv8s based on YOLOv8s. The model introduces SPPF-LSKA to enhance contextual feature extraction, employs CFC-CRB and SFC-G2 to optimize multi-scale feature fusion, and incorporates ATFLoss to improve the learning ability for minority classes and hard samples. Experimental results show that compared with the baseline model, the improved model achieves increases of 1.8\% in mAP50 and 2.1\% in mAP50-95. Compared with several mainstream detection methods, the proposed model can well meet the requirements of automatic student behavior recognition in complex classroom scenarios.
\end{abstract}

\keywords{Student classroom behavior recognition \and Object detection \and YOLOv8s  }

\section{Introduction}
In classroom teaching contexts, the instructional interactions between teachers and students generate a large amount of verbal and non-verbal behaviors, both of which significantly influence the effectiveness of classroom instruction. Consequently, research on classroom teaching behaviors has received extensive attention from scholars\cite{diadori2024nonverbal}.Regarding student classroom behavior, various scholars have conducted explorations from different perspectives, which can be broadly divided into two categories: one focuses on improving teaching effectiveness by analyzing student behaviors, while the other is dedicated to the automatic recognition and analysis of student classroom behaviors based on computer vision techniques\cite{kong2025classroom}.In the field of computer vision‑based student classroom behavior recognition, researchers have carried out rich explorations along different technical paths. First, some studies approach from the perspective of classroom learning analytics: by detecting students' facial expressions, gaze directions, and postures, they model the degree of student attention towards the teacher or the instructional content, thereby enabling quantitative analysis of classroom attention states\cite{yang2020classroom}. Such research emphasizes the correlation between behavioral information and teaching themes as well as classroom activities, providing new technical support for classroom learning analytics. Second, other studies treat student classroom state recognition as a problem of engagement classification, using transfer learning methods to identify student engagement states in real classroom environments. For example, Ikram et al \cite{ikram2023recognition}. constructed a student classroom engagement state recognition model based on VGG16, distinguishing between engaged and disengaged states in relatively less constrained real‑world classroom scenes, demonstrating the potential of deep convolutional networks in classroom behavior analysis.
Third, some research adopts multi‑module fusion or Transformer‑based detection frameworks for automatic classroom behavior recognition and analysis. For instance, Hossen et al \cite{hadash2018estimate}. integrated modules such as face detection, hand tracking, mobile phone detection, and pose estimation, and then used XGBoost to discriminate student attention states in online classrooms. Lin et al. \cite{lin2024research} based on RT‑DETR, conducted student classroom behavior detection research; by improving the backbone network, position encoding, and upsampling modules, they enhanced detection efficiency and recognition accuracy in complex classroom scenarios. These studies indicate that current classroom behavior detection is gradually moving from single‑object recognition towards multi‑feature fusion and high‑real‑time modeling.

In summary, deep learning methods have achieved certain research results in the field of classroom behavior recognition. Related studies have provided technical support for student classroom state analysis, teaching process evaluation, and smart classroom construction, also indicating that computer vision technology holds promising application prospects in educational scenarios. However, existing research still faces several urgent challenges when applied to real classroom environments. First, in classroom scenes, there are many densely distributed student targets, accompanied by large scale variations and severe mutual occlusions, which make models prone to missed detections and false detections under small‑object detection and complex background interference. Second, different student behaviors often have high visual similarity; for example, behaviors such as sitting upright, writing, and reading have limited differences in local features, leading to insufficient fine‑grained behavior discrimination capability of models. Third, classroom behavior data usually suffer from imbalanced class distribution, where some high‑frequency behaviors have abundant samples while a few behaviors have relatively insufficient samples, affecting the model's learning performance on minority classes and hard samples, thereby restricting overall recognition performance. Given these problems, it is still necessary to further optimize object detection models according to the actual characteristics of classroom teaching scenarios to improve the accuracy and robustness of student classroom behavior recognition.

To address the above deficiencies in existing research, this paper improves the YOLOv8s model for the task of student behavior recognition in real classroom environments. The main contributions are as follows.
\begin{itemize}
\item First, to cope with the problems of dense student targets, small object scales, frequent occlusions, and similar behavior categories in real classroom scenes, we construct an improved YOLOv8s‑based student classroom behavior recognition model. The model takes YOLOv8s as the base framework and makes targeted optimizations to the network architecture according to the characteristics of the classroom behavior recognition task, thereby enhancing the model's adaptability to complex classroom environments.
\item Second, we systematically improve the baseline model in three key aspects: feature extraction, feature fusion, and loss optimization. Specifically, we introduce the SPPF‑LSKA module to strengthen the network's ability to model multi‑scale contextual information. We also introduce the CFC‑CRB and SFC‑G2 modules to enhance effective fusion between shallow detail information and deep semantic information. Moreover, we adopt ATFLoss to mitigate the issues of class imbalance and hard sample learning, thus improving the detection accuracy and robustness of the model for typical classroom behaviors.
\item Third, we validate the effectiveness of the proposed method through ablation experiments and comparative experiments. Experimental results show that the improved model outperforms both the baseline model and several mainstream detection methods in terms of Precision, Recall, mAP@0.5, and mAP@0.5:0.95, indicating that the proposed method can well meet the practical requirements for automatic student classroom behavior recognition and provide technical support for intelligent classroom teaching analytics.
\end{itemize}

\section{YOLOv8 Algorithm}

YOLOv8 is an object detection model series released by the Ultralytics team on January 10, 2023, targeting multiple vision tasks such as detection and segmentation, with an emphasis on achieving a favorable balance between accuracy and inference speed~\cite{ultralytics_yolov8_docs}. Structurally, YOLOv8 continues the classic ``backbone-neck-head'' paradigm of mainstream YOLO models and introduces improvements in key modules: its backbone replaces the commonly used C3 module from earlier versions with the C2f module, which enhances representational capacity through more sufficient feature flow and gradient propagation; meanwhile, it retains the multi-scale pooling concept similar to SPPF to enlarge the receptive field, thereby improving adaptability to objects of varying sizes. The neck network typically achieves complementary cross-layer semantic and detail information through multi-scale feature fusion, enhancing detection robustness for small and occluded objects. In terms of detection head design, YOLOv8 adopts a decoupled head that separates the classification and regression branches, and no longer explicitly sets an independent objectness branch, making classification and localization learning more focused and convergence more stable. At the same time, YOLOv8 introduces an anchor-free prediction paradigm, representing bounding boxes via grid points and distance regression, which reduces reliance on prior anchor box sizes and simplifies adaptation to new datasets. Regarding training and optimization, the loss function of YOLOv8 typically consists of bounding box regression loss and classification loss, among which the regression part introduces Distribution Focal Loss to improve the precision of bounding box regression; for positive and negative sample assignment, the Ultralytics implementation provides task-aligned assignment strategies such as the Task Aligned Assigner to select positive samples more reasonably and improve the consistency between classification and localization.

From the perspective of network essence, YOLOv8 is an end-to-end object detection framework primarily based on convolutional neural networks. Its capabilities are jointly supported by basic operators, task-specific components, and joint optimization objectives, including: a feature extraction network composed of convolution and residual structures, multi-scale representation construction through cross-scale feature fusion, enhanced expressive power via nonlinear activation functions, convolution-based dense prediction heads (generating bounding boxes, classes, and confidence scores), and end-to-end gradient optimization driven by multi-task loss functions (classification, localization, and confidence loss). In addition, it relies on key components such as data augmentation, dynamic label assignment, and post-processing strategies to jointly achieve efficient detection.

\section{Proposed Algorithm}

\begin{figure}
    \centering
    \includegraphics[width=0.6\linewidth]{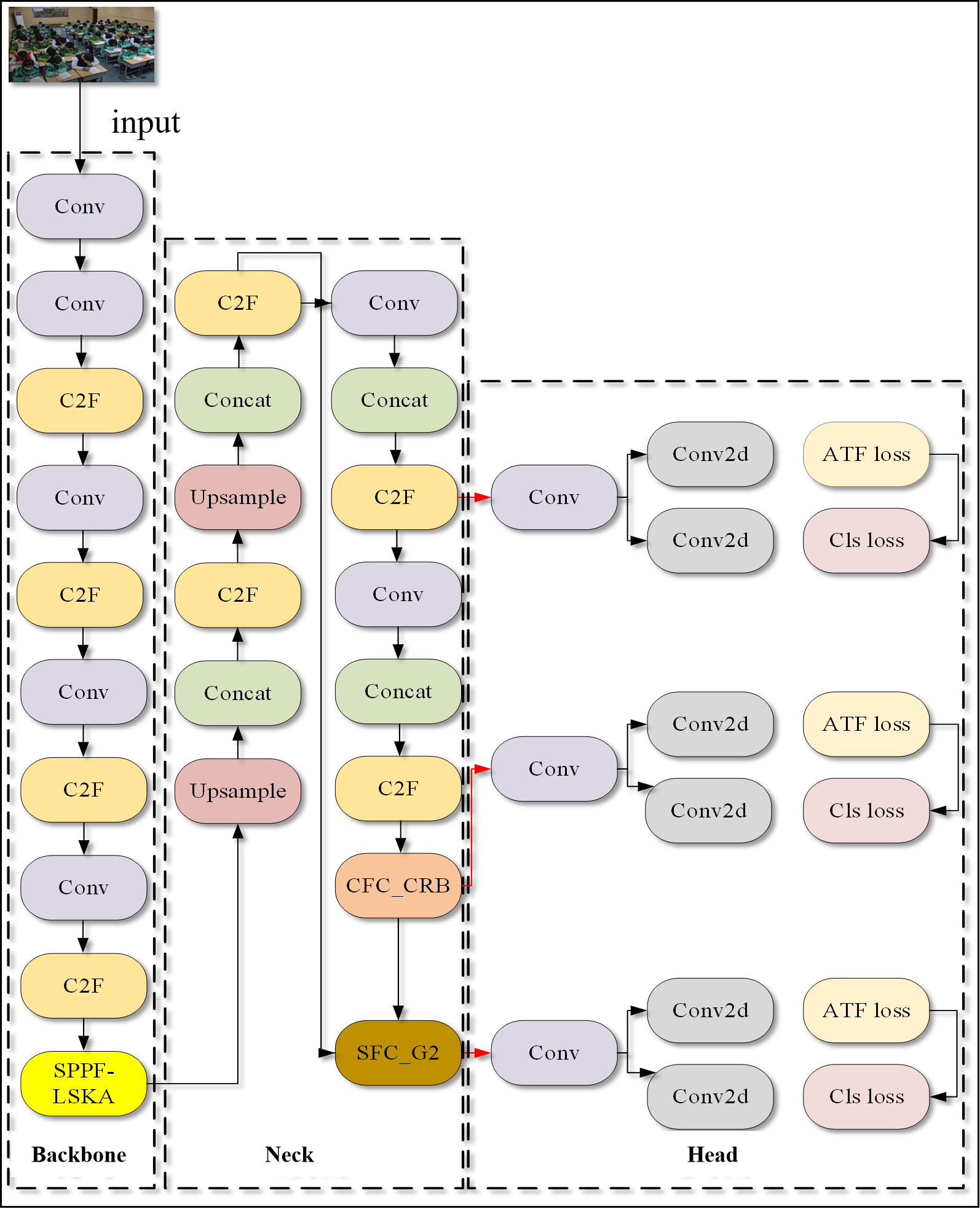}
    \caption{ALC-YOLOv8s network architecture diagram}
    \label{fig:placeholder}
\end{figure}

For the task of classroom behavior detection, the original YOLOv8s model faces three typical challenges in complex real-world classroom scenes. First, class imbalance and long-tail distributions are prevalent. Behaviors such as ``sitting upright and listening'' and ``writing'' have abundant samples, while behaviors like ``hand raising'' and ``standing'' are relatively scarce, which causes the gradient update during model training to be easily dominated by majority classes, resulting in insufficient recall for minority classes. Second, student targets in classroom images often exhibit small scales, dense distribution, and overlapping occlusions, imposing higher demands on the model's receptive field and contextual modeling capability. Third, differences among fine-grained behaviors often manifest in local regions (e.g., arm position, head orientation, torso posture), requiring more effective multi-scale feature interaction and detail preservation mechanisms. To address these issues, this paper proposes an improved model named ALC-YOLOv8s based on the YOLOv8s framework, and its network structure is shown in Figure\ref{fig:placeholder}. The improvements are specifically designed from three aspects: loss optimization, receptive field enhancement, and feature fusion, so as to enhance the model's detection robustness and fine-grained discrimination ability in complex classroom scenarios.

\subsection{SPPF-LSKA: Enlarging the Effective Receptive Field and Enhancing Spatial Contextual Representation}

\begin{figure}
    \centering
    \includegraphics[width=0.6\linewidth]{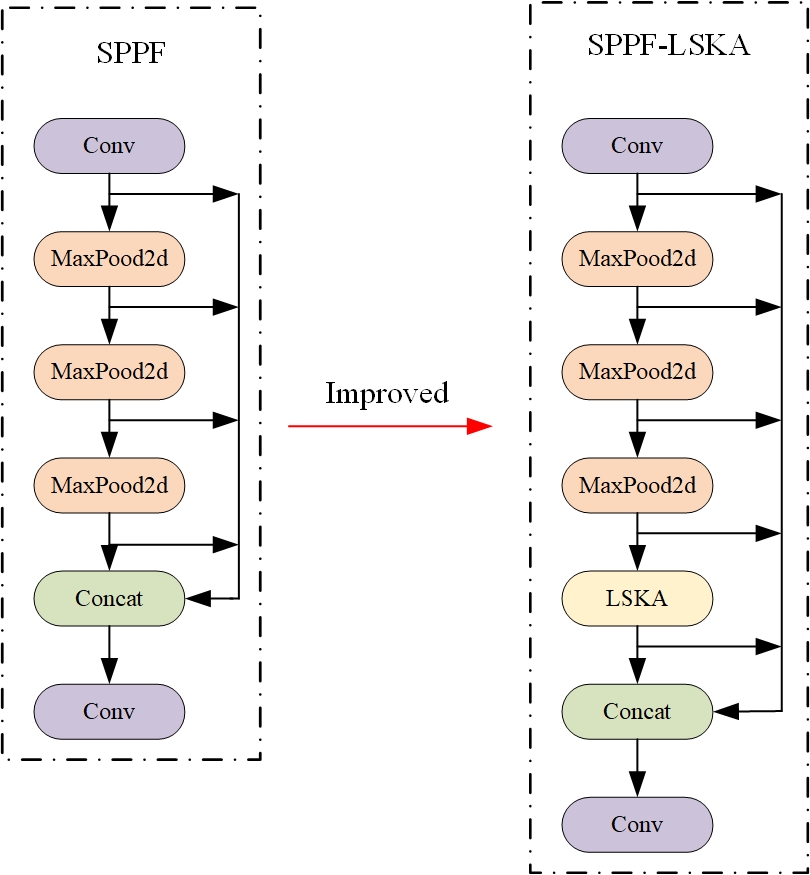}
    \caption{Schematic diagram of improvement of SPPF-LSKA module}
    \label{fig:placeholder2}
\end{figure}

To address the problems of dense targets, frequent occlusions, and a high proportion of small objects in classroom scenes, as shown in Figure \ref{fig:placeholder2}, this paper further transforms the SPPF module originally used for multi-scale context aggregation in YOLOv8s into SPPF-LSKA. Here, SPPF retains the advantage of global information aggregation brought by multi-scale pooling, while LSKA is a large-kernel separable attention mechanism that improves the model's sensitivity to target features by capturing dependencies between the spatial and channel domains~\cite{lau2024large}. Introducing the LSKA attention mechanism can compensate for the insufficient representation of structural information caused by pure pooling through stronger spatial dependency modeling. This improvement introduces richer contextual cues into the high-level semantics of the feature pyramid, enabling the model to stably locate and discriminate objects by incorporating surrounding structural information even in the presence of occlusion, background clutter, or local pose variations, thereby providing more reliable spatial evidence for distinguishing fine-grained and easily confused categories.

\subsection{CFC-CRB and SFC-G2: Enhancing Cross-layer Interaction and Fine-grained Feature Preservation}

To further improve the quality of multi-scale feature fusion, this paper introduces the CFC-CRB and SFC-G2 modules into the key paths of the backbone and neck networks to enhance cross-layer feature flow and information recalibration~\cite{li2023context}. On one hand, through more sufficient cross-layer connections and feature reorganization, the contradiction between strong semantics but insufficient details in deep layers and rich details but weak semantics in shallow layers is alleviated, thereby improving the representation capability for small targets and local actions. On the other hand, by selectively emphasizing channel/spatial information and suppressing redundancy, the discriminability and stability of feature representations are enhanced, enabling the model to retain behavior-related key cues even under complex backgrounds (e.g., blackboards, desks, book occlusions). The above modules work synergistically with SPPF-LSKA, enabling the network to obtain stronger contextual representation at the feature extraction stage and achieve higher-quality detail injection at the fusion stage, thus improving the recognition performance of fine-grained classroom behaviors.

\begin{figure}
    \centering
    \includegraphics[width=0.6\linewidth]{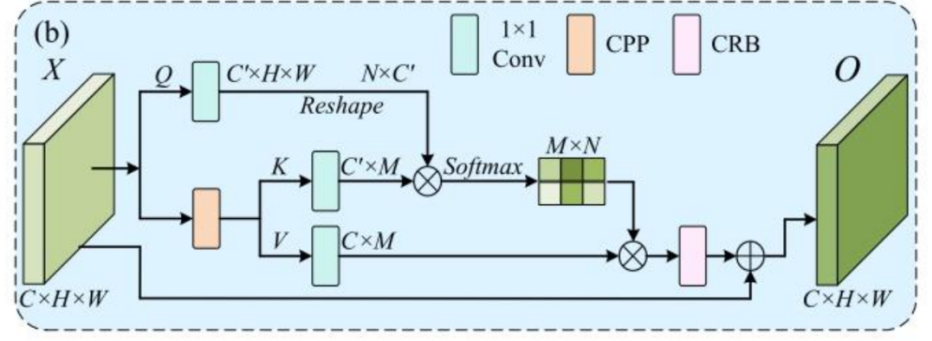}
    \caption{CFC-CRB Context Feature Calibration Module}
    \label{fig:placeholder3}
\end{figure}
As shown in Figure \ref{fig:placeholder3}, the context feature calibration module takes the input feature map as a basis and adaptively enhances spatial location features by introducing multi-scale context generated by cascaded pyramid pooling. Specifically, the input features first undergo linear mapping through one branch and are reshaped into a spatial position-wise query representation, where each pixel position corresponds to a query vector that characterizes the local semantic requirement of that position. Meanwhile, the other branch uses the contextual representation obtained from multi-scale pooling as a global prior, and generates key and value representations through linear mapping, thereby forming a context set consisting of multiple scale context units. Subsequently, the module computes the similarity between queries and keys, followed by normalization, to obtain context selection weights for each spatial position, allowing different positions to dynamically select the most relevant information from the multi-scale context and perform weighted aggregation. Through the above mechanism, the network adaptively writes multi-scale contextual information back into the feature map in a position-wise manner without destroying the spatial structure, thus improving the global consistency and scale robustness of the features.
\begin{figure}
    \centering
    \includegraphics[width=0.6\linewidth]{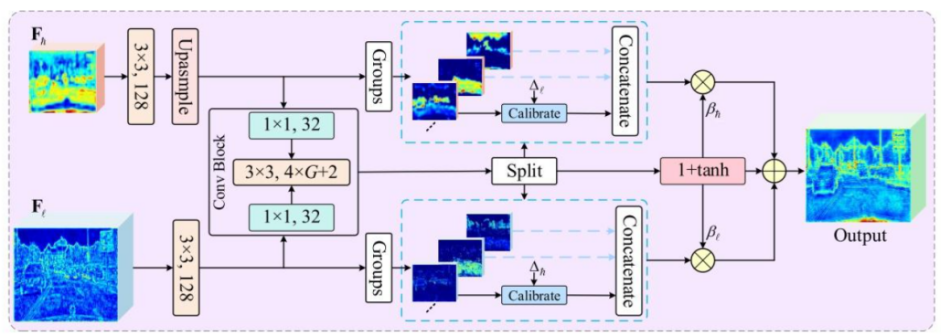}
    \caption{SFC-G2 Module Structure Diagram}
    \label{fig:placeholder4}
\end{figure}

Figure~ \ref{fig:placeholder4} illustrates the structure of the SFC-G2 module. This spatial feature calibration module takes high-level features and low-level features as inputs. First, it unifies the number of channels through convolution and upsamples the high-level features to align the spatial resolution, providing a consistent spatial coordinate system for cross-layer fusion. Then, the module employs lightweight convolutional blocks to generate displacement information for spatial calibration, and divides the two feature streams into groups to reduce computational cost and enhance the specificity of local modeling. Within each group, the predicted displacement is used to resample and align the features, achieving spatial offset correction and detail compensation for key regions. After that, the calibration results from all groups are concatenated along the channel dimension to form two enhanced representations. To avoid semantic conflicts caused by simple addition, the module further introduces a gating weight to adaptively fuse the two calibrated feature streams via weighted fusion, where the weight magnitude is constrained through nonlinear mapping to stabilize training. The final output is formed by combining the weighted fusion result with the original information, enabling the network to strengthen low-level edge and texture details while maintaining high-level semantic consistency, thereby improving the representational capacity for object location and structure.

\subsection{ATFLoss: Alleviating Class Imbalance and Enhancing Hard Example Learning}

To address the training bias caused by class imbalance, this paper replaces the classification learning strategy in the original training objective with Adaptive Threshold Focal Loss (ATFLoss)~\cite{yang2024eflnet}. This loss applies differentiated weights to samples of different categories and difficulties, so that the model no longer merely follows the dominant gradient from ``easy-to-classify, majority-class'' samples during training, but instead maintains higher optimization attention on ``minority-class, hard, and easily confused samples''. Specifically, ATFLoss introduces an adaptive modulation mechanism based on the classification error, where the gradient contribution of high-confidence easy samples is appropriately suppressed, while that of low-confidence or hard-to-distinguish samples is enhanced, thereby increasing the effective learning intensity of minority-class behaviors (e.g., hand raising, discussion) in the later stages of training. This strategy improves the model's class discrimination boundary without altering the inference structure, reduces the phenomenon of ``overfitting on majority classes and underfitting on minority classes'', and thus enhances overall detection performance, especially the recall and F1 score of few-shot categories.

In summary, ALC-YOLOv8s, through the combined improvements of ``ATFLoss (training objective layer) + SPPF-LSKA (context and receptive field layer) + CFC-CRB/SFC-G2 (feature fusion layer)'', achieves targeted enhancements for key pain points of classroom behavior detection while maintaining the lightweight and deployment efficiency advantages of YOLOv8s as much as possible. In subsequent experiments, this paper will validate the contribution of each improved component through ablation studies, and systematically evaluate the effectiveness and generalization ability of ALC-YOLOv8s based on metrics including mAP, recall, and changes in minority-class indicators.

\section{Experiments and Analysis}

\subsection{Dataset}

The data used in this paper are derived from a self-constructed annotated dataset. This study selected classroom video recordings of physics and other science subjects from public course resources. In addition, physics classroom recordings were obtained with the consent of a university and the students involved. Seven common categories of student classroom behaviors are defined in this study: sitting upright and listening, looking down (distracted), looking around, reading, writing, standing, and hand raising. The detailed descriptions are shown in Table~\ref{tab:behavior_categories}.

\begin{table}[htbp]
\centering
\caption{Student classroom behavior categories and descriptions}
\label{tab:behavior_categories}
\begin{tabular}{|l|p{10cm}|}
\hline
\textbf{Behavior} & \textbf{Description} \\
\hline
Sitting upright and listening & Body upright, facing forward, eyes focused on the teaching subject, with timely non-verbal feedback, no unnecessary small movements, and ability to respond to the teacher promptly. \\
\hline
Looking down (distracted) & Head continuously lowered, gaze deviating from the teaching area, disengaged from classroom content, showing signs of daydreaming, boredom, or immersion in personal thoughts. \\
\hline
Looking around & Frequent aimless head turning, gaze wandering to the environment or others, often accompanied by whispering, passing objects, and other irrelevant social behaviors. \\
\hline
Reading & Holding textual materials, eyes steadily focused on the content, synchronized with instruction (e.g., following along, previewing, reviewing), reflecting active information processing. \\
\hline
Writing & Holding a pen to write or answer on a notebook, including note-taking, copying from the blackboard, doing exercises, drawing, etc., highly related to the lecture content. \\
\hline
Standing & Changing from sitting to standing posture when responding in class and answering the teacher's questions. \\
\hline
Hand raising & Raising one or both arms, elbows off the desk, palms facing forward or holding a pen, indicating a desire to speak, ask questions, or seek help, reflecting participation confidence. \\
\hline
\end{tabular}
\end{table}

\subsection{Environment}

The hardware and software configurations used in the experiments are listed in Table~\ref{tab:environment}.

\begin{table}[htbp]
\centering
\caption{Hardware and software environment}
\label{tab:environment}
\begin{tabular}{|l|l|}
\hline
\textbf{Item} & \textbf{Configuration} \\
\hline
Deep learning framework & PyTorch 1.11.0 \\
\hline
Programming language & Python 3.8 \\
\hline
Operating system & Ubuntu 20.04 \\
\hline
CUDA version & CUDA 11.3 \\
\hline
GPU & RTX 4090 (24GB) × 1 \\
\hline
CPU & 16 vCPU Intel(R) Xeon(R) Platinum 8470Q \\
\hline
\end{tabular}
\end{table}

\subsection{Ablation Experiments}

As shown in Table~\ref{tab:ablation}, this paper conducts ablation experiments based on the YOLOv8s baseline model around three key improvements: A (SPPF-LSKA module), B (feature fusion enhancement structure with CFC-CRB and SFC-G2), and C (ATFLoss loss function). The experimental results show that introducing each improvement individually brings stable performance gains, with consistent improvements across Precision, Recall, mAP50, and mAP50-95. Specifically, YOLOv8s+A increases mAP50 and mAP50-95 to 0.865 and 0.720, respectively; YOLOv8s+B further improves them to 0.866 and 0.721; YOLOv8s+C (with ATFLoss) achieves more significant improvements (mAP50 = 0.870, mAP50-95 = 0.725), indicating that optimizing hard examples and bounding box regression quality at the loss level contributes more directly to overall detection accuracy. Furthermore, pairwise combination experiments show certain complementarity and additive effects among the components. Among them, the B+C combination achieves the best dual-module performance (mAP50 = 0.874, mAP50-95 = 0.732), reflecting that the enhancement of feature representation/fusion capability together with the hard example emphasis in the training objective can synergistically improve the model's discriminative ability for multi-scale targets and easily confused categories. Finally, the full model YOLOv8s+A+B+C achieves the best performance (P = 0.826, R = 0.831, mAP50 = 0.879, mAP50-95 = 0.736), which is 0.011, 0.005, 0.018, and 0.021 higher than the baseline, respectively. This verifies that the proposed improvements form an effective chain of “feature extraction – feature fusion – optimization objective”, thereby significantly improving overall detection performance and generalization ability.

\begin{table}[htbp]
\centering
\caption{Ablation experiment results of each module in the ALC-YOLOv8s model}
\label{tab:ablation}
\begin{tabular}{|l|c|c|c|c|}
\hline
\textbf{Model} & \textbf{P} & \textbf{R} & \textbf{mAP50} & \textbf{mAP50-95} \\
\hline
YOLOv8s     & 0.815 & 0.826 & 0.861 & 0.715 \\
YOLOv8s+A   & 0.818 & 0.827 & 0.865 & 0.720 \\
YOLOv8s+B   & 0.820 & 0.828 & 0.866 & 0.721 \\
YOLOv8s+C   & 0.823 & 0.829 & 0.870 & 0.725 \\
YOLOv8s+A+B & 0.821 & 0.829 & 0.868 & 0.723 \\
YOLOv8s+A+C & 0.824 & 0.830 & 0.872 & 0.729 \\
YOLOv8s+B+C & 0.825 & 0.830 & 0.874 & 0.732 \\
YOLOv8s+A+B+C & 0.826 & 0.831 & 0.879 & 0.736 \\
\hline
\end{tabular}
\end{table}

\subsection{Comparative Experiments}

Under the same dataset and evaluation protocol, comparative experiments are conducted on SSD, RT-DETR, Faster R-CNN, YOLOv5s, YOLOv8s, YOLOv11s, and the proposed ALC-YOLOv8s. The results are shown in Table~\ref{tab:comparison}. Overall, ALC-YOLOv8s achieves the best comprehensive detection performance, indicating that the introduced structural improvements significantly enhance the overall localization and classification ability under multi-threshold conditions while maintaining high precision and recall. Compared with YOLOv5s, YOLOv8s, and YOLOv11s, all metrics are uniformly improved, demonstrating that the proposed method still achieves considerable gains on strong baseline models. The two-stage method Faster R-CNN achieves an mAP50 of 0.822 but only 0.577 in mAP50-95, suggesting its relatively insufficient fine localization capability under high IoU thresholds. The lightweight SSD has the lowest overall metrics, reflecting its limited adaptability to scale variation and occlusion in complex scenes. In summary, ALC-YOLOv8s achieves leading performance in the precision-recall trade-off and multi-threshold average precision, validating the effectiveness and generalization potential of the proposed improvement strategies.

\begin{table}[htbp]
\centering
\caption{Comparative experiment results of object detection algorithms}
\label{tab:comparison}
\begin{tabular}{|l|c|c|c|c|}
\hline
\textbf{Model} & \textbf{P} & \textbf{R} & \textbf{mAP50} & \textbf{mAP50-95} \\
\hline
SSD           & 0.611 & 0.637 & 0.684 & 0.471 \\
RT-DETR       & 0.745 & 0.763 & 0.773 & 0.607 \\
Faster R-CNN  & 0.761 & 0.792 & 0.822 & 0.577 \\
YOLOv5s       & 0.823 & 0.827 & 0.860 & 0.698 \\
YOLOv8s       & 0.815 & 0.826 & 0.861 & 0.715 \\
YOLOv11s      & 0.806 & 0.834 & 0.870 & 0.719 \\
ALC-YOLOv8s   & 0.826 & 0.831 & 0.879 & 0.736 \\
\hline
\end{tabular}
\end{table}

\section{Conclusion}

Focusing on the problem of student behavior recognition in complex classroom environments, this paper proposes the ALC-YOLOv8s model by introducing targeted improvements to YOLOv8s. The model enhances recognition capability for small targets, occluded objects, and similar behavior categories through three aspects: receptive field enhancement, feature fusion optimization, and loss function improvement. Experimental results show that the proposed method achieves favorable performance across multiple evaluation metrics, verifying the effectiveness of the model design and demonstrating that the improved network has practical application value for student classroom behavior recognition tasks. Overall, this study provides technical support for automatic classroom behavior analysis. However, there remains room for further improvement in data scale, scene diversity, and cross-scene generalization. Future work can continue in-depth research by incorporating temporal information and more teaching scenarios.

\bibliographystyle{unsrt}  
\bibliography{references}

@article{hadash2018estimate,
  title={Estimate and Replace: A Novel Approach to Integrating Deep Neural Networks with Existing Applications},
  author={Hadash, Guy and Kermany, Einat and Carmeli, Boaz and Lavi, Ofer and Kour, George and Jacovi, Alon},
  journal={arXiv preprint arXiv:1804.09028},
  year={2018}
}

@article{diadori2024nonverbal,
  title={Nonverbal communication in classroom interaction and its role in Italian foreign language teaching and learning},
  author={Diadori, Pierangela},
  journal={Languages},
  volume={9},
  number={5},
  pages={164},
  year={2024},
  publisher={MDPI}
}

@article{kong2025classroom,
  title={Classroom behavior analysis and digital teaching quality evaluation based on spatiotemporal graph neural network},
  author={Kong, Yang and Dong, Rongwei and Zhang, Hui},
  journal={Discover Artificial Intelligence},
  volume={5},
  number={1},
  pages={404},
  year={2025},
  publisher={Springer}
}

@article{lin2024research,
  title={Research on student classroom behavior detection based on the real-time detection transformer algorithm},
  author={Lin, Lihua and Yang, Haodong and Xu, Qingchuan and Xue, Yanan and Li, Dan},
  journal={Applied Sciences},
  volume={14},
  number={14},
  pages={6153},
  year={2024},
  publisher={MDPI}
}

@article{yang2020classroom,
  title={In-classroom learning analytics based on student behavior, topic and teaching characteristic mining},
  author={Yang, Bohong and Yao, Zeping and Lu, Hong and Zhou, Yaqian and Xu, Jinkai},
  journal={Pattern Recognition Letters},
  volume={129},
  pages={224--231},
  year={2020},
  publisher={Elsevier}
}

@article{ikram2023recognition,
  title={Recognition of student engagement state in a classroom environment using deep and efficient transfer learning algorithm},
  author={Ikram, Sana and Ahmad, Haseeb and Mahmood, Nasir and Faisal, CM Nadeem and Abbas, Qaisar and Qureshi, Imran and Hussain, Ayyaz},
  journal={Applied Sciences},
  volume={13},
  number={15},
  pages={8637},
  year={2023},
  publisher={MDPI}
}

@misc{ultralytics_yolov8_docs,
  title        = {Ultralytics YOLOv8 Documentation},
  author       = {Ultralytics},
  year         = {2023},
  url          = {https://docs.ultralytics.com/},
  note         = {Accessed: 2026-04-07}
}

@article{lau2024large,
  title={Large separable kernel attention: Rethinking the large kernel attention design in cnn},
  author={Lau, Kin Wai and Po, Lai-Man and Rehman, Yasar Abbas Ur},
  journal={Expert Systems with Applications},
  volume={236},
  pages={121352},
  year={2024},
  publisher={Elsevier}
}

@article{li2023context,
  title={Context and spatial feature calibration for real-time semantic segmentation},
  author={Li, Kaige and Geng, Qichuan and Wan, Maoxian and Cao, Xiaochun and Zhou, Zhong},
  journal={IEEE Transactions on Image Processing},
  volume={32},
  pages={5465--5477},
  year={2023},
  publisher={IEEE}
}

@article{yang2024eflnet,
  title={EFLNet: Enhancing feature learning network for infrared small target detection},
  author={Yang, Bo and Zhang, Xinyu and Zhang, Jian and Luo, Jun and Zhou, Mingliang and Pi, Yangjun},
  journal={IEEE Transactions on Geoscience and Remote Sensing},
  volume={62},
  pages={1--11},
  year={2024},
  publisher={IEEE}
}

\end{document}